\documentclass[wcp]{jmlr}

\usepackage{longtable}

\usepackage{booktabs}

\usepackage{lineno}
\usepackage{diagbox}

\makeatletter
\let\Ginclude@graphics\@org@Ginclude@graphics 
\makeatother

\jmlrvolume{304,}
\jmlryear{2025}
\jmlrworkshop{ACML 2025}

\title[DiffGap]{Bridging the Gap between Learning and Inference for Diffusion-Based Molecule Generation}

\author{\Name{Peidong Liu} \Email{peidong\_liu@stu.scu.edu.cn}\\
    \Name{Wenbo Zhang} \Email{zhangwenbo01@xidian.edu.cn}\\
    \Name{Wei Ju} \Email{juwei@scu.edu.cn}\\
    \Name{Jiancheng Lv} \Email{lvjiancheng@scu.edu.cn}\\
    \Name{Xianggen Liu} \Email{liuxianggen@scu.edu.cn}\\
\addr{Sichuan University
No. 24 South Section 1, Yihuan Road
Chengdu, Sichuan 610065, China}
}

\editors{Hung-yi Lee and Tongliang Liu}

\begin{document}
\makeatletter
\let \@jmlrpages \@empty
\makeatother

\maketitle

\begin{abstract}
The paradigm shift toward structure-driven molecule generation has been propelled by advances in deep generative models, such as variational auto-encoders and diffusion models. However, these generative models for molecular design remain constrained by exposure bias, error accumulation, and suboptimal handling of activity cliffs. Here, we introduce DiffGap, a diffusion-based framework that integrates adaptive sampling and pseudo-molecule estimation to bridge the gap between training objectives and inference dynamics in 3D molecule generation. By dynamically aligning intermediate denoising steps with realistic generation trajectories, DiffGap enables the diffusion model to adapt to input biases in advance during the training phase. A temperature annealing module further controls the aligning strength of the adaptive alignment process, ensuring stable learning of the data distribution. Evaluated on the CrossDocked2020 benchmark, DiffGap outperforms existing methods in docking scores and binding affinity, demonstrating superior fidelity in generating drug-like molecules. Our work establishes a principled approach to harmonize generative training with inference mechanics, offering a robust computational toolkit for accelerating structure-based therapeutic discovery.
The source code of DiffGap is available at https://github.com/neusymlab/DiffGap.
\end{abstract}

\section{Introduction}
The pursuit of targeted therapeutic agents represents a cornerstone of modern pharmaceutical research, where molecular design is guided by precise three-dimensional interactions between ligands and disease-associated proteins \citep{sneader2005drug}. This paradigm shift from serendipitous discovery to structure-driven design has been accelerated by advancements in structural biology \citep{ batool2019structure, liu2021deep} and computer-aided rational design\citep{mandal2009rational,alphafold, alphafold3}. In particular, structure-based drug design typically employs molecular docking simulations, pharmacophore modeling, and free energy perturbation calculations to virtual screen promising compounds from curated chemical libraries. While virtual screening \citep{mayr2009novel, zhang2022molecular} remains prevalent, its efficacy is fundamentally constrained by the combinatorial explosion of drug-like chemical space ($\sim10^{60}$ potential molecules) \citep{bohacek1996art}.

Generative approaches have emerged as promising alternatives to exhaustive screening. Deep generative models (DGMs) attempt to navigate chemical space to generate 2D molecules from learned latent representations \citep{jin2018junction}, yet their performance degrades when target properties diverge from training data distributions \citep{ning2023input}. Combinatorial optimization methods \citep{you2018graph, jensen2019graph, paassen2022recursive} show superior potential but remain underexplored for target-specific design. Recent attempts using simulated annealing \citep{xue2025targetsa} demonstrate limitations in handling activity cliffs---abrupt property changes from minor structural modifications \citep{pemasinghe2021simulated} due to their myopic optimization strategies.

The integration of geometric deep learning has revolutionized molecular representation learning. Early 2D approaches \citep{jin2018junction, walters2020applications, lim2020scaffold} gave way to 3D-equivariant architectures \citep{fuchs2020se, satorras2021En} that preserve rotational and translational symmetries critical for molecular geometric structures. Diffusion models leveraging 3D-equivariant networks \citep{targetdiff, guan2023decompdiff, huang2024binddm} to gradually perform the de-noising of the geometric topological structures, achieving state-of-the-art performance in 3D molecule generation. However, the iterative denoising process (typically requiring $\geq$1,000 steps) introduces error accumulation and exposure bias between training objectives and inference conditions analogous to autoregressive sequence generation \citep{ning2023elucidating}.

However, the success of these diffusion models masks a critical vulnerability rooted in their iterative generation process. This vulnerability stems from a fundamental discrepancy between how the models are trained and how they perform inference, and is known as exposure bias. During training, the model always learns to denoise a perfectly conditioned state $M_t$ to a pristine, ground-truth molecule $M_0$. In contrast, during inference, the model must denoise a state $\hat{M}_t$ which is the result of its own prediction from the previous step, $\hat{M}_{t+1}$. The model is therefore never exposed to its own errors during training, making it fragile when faced with them when autoregressive-style generation \citep{ning2023elucidating}.

To address these limitations, we present DiffGap, a novel diffusion framework incorporating adaptive sampling through pseudo-molecule estimation. Our key insight of the adaptive sampling lies in dynamically aligning the training inputs with the realistic generation trajectories, rather than following teacher-forcing denoising computations. This approach reduces the train-inference discrepancy by treating intermediate predictions as conditional inputs for subsequent steps. We also introduce a temperature annealing module to control the aligning strength of the adaptive sampling process. Extensive evaluations on the CrossDocked2020 benchmark \citep{crossdock2020} demonstrate DiffGap's superiority in generating molecules with optimized binding affinities and 3D complementarity.

In summary, our principal contributions are threefold:
\begin{itemize}
    \item We propose a generative framework, called DiffGap, to address the exposure bias issue of diffusion models by an adaptive sampling strategy.
    \item The DiffGap framework introduces the construction strategy of pseudo-molecules for training, which is an optimal estimation of the input condition to mimic the inference computations.
    \item Empirical results show that the molecules generated by DiffGap achieve state-of-the-art docking scores and superior quality on the binding affinity.
\end{itemize}

\section{Related work}
\subsection{Molecule generation}
Existing molecular generation models can be categorized into four groups: string-based, image-based, 2D graph-based, and 3D structure-based. The most common molecular string representation is SMILES~\citep{weininger1988smiles}, where researchers can reuse language models like RNN and Transformer and quickly apply them to molecular generation tasks following the text approach~\citep{schoenmaker2023uncorrupt,brahmavar2024generating}. For example, researchers trained RNN and its variants on randomized SMILES strings to improve the uniqueness of generated molecules~\citep{grisoni2020bidirectional,arus2019randomized} and ChatMol empowered large language model for conversational molecular design diagram~\citep{zeng2024chatmol}.
2D molecular image representations~\citep{walters2020applications} and 2D graph representations~\citep{lim2020scaffold, jin2018junction} employed CNNs and GNNs respectively for more atom connection information than string representations.
Recently, structure-based models like GraphBP, Pocket2Mol, and diffusion models took the 3D structure and equivariant properties of molecules into account, showing advantages in molecular affinity~\citep{GraphBP, peng2022pocket2mol, targetdiff, guan2023decompdiff, huang2024binddm}. Our work follows the same denoising theory of diffusion models, but adopts a new training framework (i.e., the adaptive sampling strategy) to improve the generation quality.

\subsection{Diffusion models}
Introduced by ~\citep{sohl2015dpm} and developed by ~\citep{ddpm, song2020score}, diffusion models have been applied in various fields like unconditional image generation~\citep{ddpm}, text-to-image generation~\citep{nichol2021glide}.
Recently, diffusion models have also been applied to molecular generation tasks, particularly in the field of Structure-Based Drug Design (SBDD). For instance, TargetDiff~\citep{targetdiff} combined with an SE(3)-equivalent network, has surpassed the previous SOTA method, Pocket2Mol~\citep{peng2022pocket2mol}, with a significant docking score on the CrossDock2020 dataset. 
BindDM adaptively extracted subcomplex and captured the protein-ligand interactions exactly to obtain higher docking affinity than the above models~\citep{huang2024binddm}. The aforementioned models primarily focus on adapting molecular data and properties, proposing various strategies to process such inputs. In contrast, our work centers on the universal sampling strategy of the diffusion models for molecule generation, which is effective for multiple diffusion-based methods, as validated by the empirical results.

\subsection{Exposure bias}
Exposure bias has been widely studied in sequence generation tasks, particularly in natural language processing (NLP) applications and recommendation algorithms~\citep{yang2018unbiased, lamb2016professor,zhang2019bridge}. The term exposure bias refers to the discrepancy between training and inference conditions in sequence models. During training, models are conditioned on ground truth data (teacher forcing), while during inference, they generate sequences based on previous predictions, leading to error accumulation~\citep{lamb2016professor}. 
Diffusion models share the same problem, but are few solutions~\citep{ning2023input,ning2023elucidating}.
To complement research in this domain, we provide a precise prediction of the input condition to narrow the input gap between training and generation.
Different from it, our method provides a more precise prediction of the input condition by Bayesian estimation to narrow the input gap between training and generation.

\section{Problem definition}\label{subsec:def}

The problem of structure-based 3D molecular generation is defined as a conditional generation process under the specific protein pocket. 
Formally speaking, a data point consists of pairs of proteins $\mathcal{P}$ and molecular conformations $\mathcal{M}$. The molecular conformation is represented by the concatenation of its atomic 3D Cartesian coordinates $x\in\mathcal{R}^{m\times 3}$ and one-hot encoded atomic types $v\in\mathcal{R}^{m\times k}$ ($m$ denotes the number of atoms in a molecule and $k$ shows the number of potential atomic types), and so is the protein.
That is, the goal of structure-based 3D molecular generation is to generate a reasonable molecular conformation $\mathcal{M}=[x,v]$ given the protein $\mathcal{P}=[x_\mathcal{P},v_\mathcal{P}]$ using a novel diffusion model with less bias.

\section{Methodology}\label{sec:method}

\subsection{Classic diffusion process}\label{subsec:cdp}
The diffusion model defines the \textit{diffusion process} with the data $\mathbf{x}_0 \sim q(\mathbf{x}_0)$, which is a Markov chain that incrementally adds Gaussian noise in equation~\ref{cdp:eq:fwd-os} using schedule hyperparameters $\beta_1,\dots,\beta_T$.
\begin{align}
q(\mathbf{x}_{1:T}|\mathbf{x}_0)=\prod^T_{t=1}q(\mathbf{x}_{t}|\mathbf{x}_{t-1})
,\quad\quad\quad\quad
q(\mathbf{x}_{t}|\mathbf{x}_{t-1}) = \mathcal N(\mathbf{x}_t;\sqrt{1-\beta_t}\mathbf{x}_{t-1},\beta_t\mathbf I)\label{cdp:eq:fwd-os}
\end{align}
Based on the properties of the Gaussian distribution, this process of incrementally adding noise $q(\mathbf{x}_t|\mathbf{x}_0)$ can be simplified in equation~\ref{cdp:eq:fwd} with notations $\alpha_t=1-\beta_t;\;\bar\alpha_t=\prod^t_{s=1}\alpha_s$.
\begin{equation}
  \label{cdp:eq:fwd}
  q(\mathbf{x}_t|\mathbf{x}_0)=\mathcal N(\mathbf{x}_t;\sqrt{\bar\alpha_t}\mathbf{x}_0, (1-\bar\alpha_t)\mathbf I)
\end{equation}

The diffusion model is a parameterized Markov chain that models a latent variable model of the form $p_\theta(\mathbf{x}_0)=\int p_\theta(\mathbf{x}_{0:T})d\mathbf{x}_{1:T}$, used to learn the reverse Gaussian denoising process of the diffusion process $p_\theta(\mathbf{x}_{t-1}|\mathbf{x}_t)$, i.e., the {reverse process}. The {reverse process} can be formalized as a normal distribution in equation~\ref{cdp:eq:reverse} because the forward process consists of thousands of steps and each one 
$q(\mathbf{x}_t|\mathbf{x}_{t-1})$ follows a Gaussian distribution.
Consequently, we can utilize $\mu_\theta(\mathbf{x}_t,t)$ and $\Sigma_\theta(\mathbf{x}_t,t)$ to denote the parameters of the normal distribution for a single-step reverse process under the neural network.
\begin{align}
p_\theta(\mathbf{x}_{0:T})=p(\mathbf{x}_T)\prod^T_{t=1}p_\theta(\mathbf{x}_{t-1}|\mathbf{x}_t)
\label{cdp:eq:reverse}
,\quad\quad\quad\quad
p_\theta(\mathbf{x}_{t-1}|\mathbf{x}_t) = \mathcal N(\mathbf{x}_{t-1};\mu_\theta(\mathbf{x}_t,t),\Sigma_\theta(\mathbf{x}_t,t))
\end{align}

Intuitively, what we need to know is the denoising distribution $q(\mathbf{x}_{t-1}|\mathbf{x}_t)$ of the data, but it is not tractable. Instead, we can compute the posterior probability of the data point $q(\mathbf{x}_{t-1}|\mathbf{x}_t,\mathbf{x}_0)$ given the original data point as in equation~\ref{cdp:eq:bayes}.
\begin{align}\label{cdp:eq:bayes}
q(\mathbf{x}_{t-1}|\mathbf{x}_t,\mathbf{x}_0)=\mathcal{N}(x_{t-1};\tilde\mu_t(\mathbf{x}_t,\mathbf{x}_0),\tilde\beta\mathbf{I})
\end{align}
In this way, the ground truth of the denoising probability can be accurately approximated.

\subsection{DiffGap framework}\label{subsec:pdm}

DiffGap adopts the diffusion process to learn the conformations $M$ of a molecule in the 3D space, where the molecule state $M_t$ at step $t$ can be determined by the previous state $M_{t-1}$  and the protein structure $\mathcal{P}$. 
\begin{align}
q(M_t|M_{t-1},\mathcal{P})= P(x_t|x_{t-1}, \mathcal{P})P(v_t|v_{t-1}, \mathcal{P}) \label{css:eq:fwd}
\end{align}
where $M_t=[x_t,v_t]$ is the molecule state at step $t$. To reduce the computational complexity, we assume the atom coordinates $x_t$ and the identity $ v_t$ are independent in the estimation of the transformation probability $q(M_t|M_{t-1},\mathcal{P})$. Similar to equation~\ref{cdp:eq:fwd-os}, we model the transformation probability of atomic positions $ P(x_t|x_{t-1}, \mathcal{P}) $ and that of the atomic types $P(v_t|v_{t-1}, \mathcal{P})$ using a normal distribution and a categorical distribution ($K$ represents the number of categories), respectively.

The objective of diffusion models is to narrow the divergence between the denoising Gaussian distribution (i.e., ground truth) $q(M_{t-1}|M_{t}, M_0, \mathcal{P} )$ and the predicted distribution $p_\theta(M_{t-1}|M_{t},\mathcal{P})$, where the condition $M_{t}$ is sampled from the unbiased Gaussian distribution. 

\begin{align}
\label{cdp:eq:loss-step}
L_{t-1} &= \mathbb{E}_q\left[\sum_{t\ge1}D_{\rm KL}(q(M_{t-1}|M_{t},M_0, \mathcal{P} )||p_\theta(M_{t-1}|M_{t},\mathcal{P}))\right]\\
&=\mathbb{E}_q\left[
    \frac{1}{\Sigma_\theta(M_t,t)}||\tilde\mu_t(M_t, M_0)-\mu_\theta(M_t,t)||^2
\right]+C
\end{align}
where $C$ is a constant and is independent of the neural network.

\begin{figure*}[h]
  \centering
  \includegraphics[width=\linewidth]{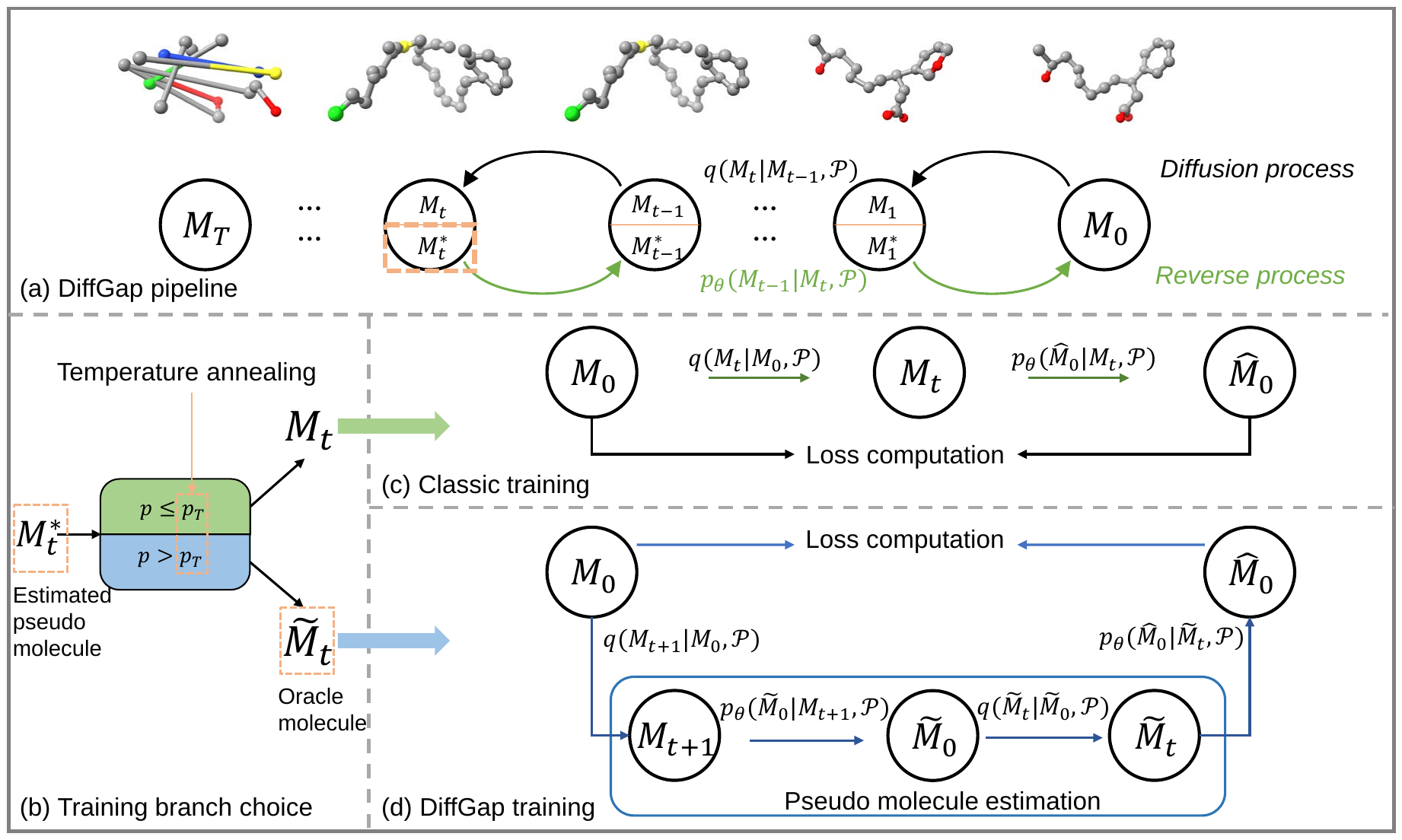}
  \caption{The overview of DiffGap pipeline under target-aware molecule generation task. DiffGap is consistent with the logic of DDPM, with the difference lying in the reverse process during training, shown in Figure (a). In the reverse process of training, the ground truth is selected probabilistically between the original ground truth (denoted as $M_i$ like $M_t$) and the model's real-time predicted value (denoted as $\widetilde{M}_i$ like $\widetilde M_t$), with a probability $p_T$ favoring the original value in Figure (b). What's more, the probability $p_T$ is periodically updated by temperature annealing. Figures (c) and (d) display the two training pathways guided by $p_T$ and we will discuss it further in section~\ref{sec:method}.
  }
\label{css:fig:DiffGap}
\end{figure*}

As for the inference of DiffGap, it starts with Gaussian noises $M_T$ and iteratively denoises from the previous result, finally achieving the goal. That is,
\begin{align}
p_\theta(\hat{M}_{0:T}|\mathcal{P})=p(\hat{M}_T)\prod^T_{t=1} \overbrace{p_\theta(\hat{M}_{t-1}|\hat{M}_t,\mathcal{P}) }^{\text{Condition varies from equation~\ref{cdp:eq:loss-step}.}}
,\quad\quad\quad\quad
\hat{M}_t\sim p_\theta(\hat{M}_{t}|\hat{M}_{t+1},\mathcal{P})
\end{align}
where the condition $\hat{M}_t$ of the denoising process is generated in the last step $p_\theta(M_t|M_{t+1})$, instead of the sample from the unbiased Gaussian distribution.

However, the above inference process of DiffGap will suffer serious exposure bias due to the discrepancy between training and inference. In 3D molecule generation, the conformation space of 3D molecules is huge and rough, with most breaking the chemical and physical rules. A small atom-type shift could result in an unrealistic molecule. During inference, the reverse process can be viewed as an autoregressive generative process iterated multiple times in time steps, generating $M_0$ by gradually denoising $M_t$ sampled randomly according to equation~\ref{cdp:eq:reverse}. However, general diffusion models require a large number of iteration steps (typically 1000), leading to error accumulation and exposure bias issues.



\textbf{Adaptive sampling strategy.}\label{subsec:css}
To address the exposure bias issue in traditional diffusion models, we propose an adaptive sampling strategy, which reduces the discrepancy in data distribution between training and inference by introducing reasonable noise into the training phase.
The core idea of DiffGap is to utilize model-predicted conformations as ground truth during training probabilistically.
In particular, we denote $M^*_t$ as a perturbed condition
in the denoising probability $q(M_{t-1}|M_t, M_0)$ to replace the ground truth sample $M_t$, which is unavailable in the inference phase. Therefore, the objective function of DiffGap is
\begin{align}
L_{t-1} = \mathbb{E}_q\left[\sum_{t\ge1}D_{\rm KL}(q(M_{t-1}|M^*_t,M_0, \mathcal{P} )||p_\theta(M_{t-1}|M^*_t,\mathcal{P}))\right]
,\quad
M^*_t = {\rm Perturbation}(M_t)
\label{cdp:eq:loss-m}
\end{align}

Since molecule $M_t=[x_t,v_t]$ has two components, we need to compute the atom coordinate loss and the atom type loss respectively using equation~\ref{cdp:eq:loss-step}. We measure the distance between the predicted 
 atom coordinates and the ground truth for coordinate loss, and take the KL-divergence of categorical distributions (i.e., $c(v_t,v_0)$) for the atom type loss. Then the objective function can be further transformed to
\begin{align}
L_{t-1} = L_{t-1}^{(x)}+\lambda L_{t-1}^{(v)}
,\quad\quad
L^{(x)}_{t-1} = \gamma_t||x_0-\hat{x}_0||^2+C
,\quad\quad
L^{(v)}_{t-1} = \sum_kc(v_t,v_0)_k{\rm log}\frac{c(v_t,v_0)_k}{c(v_t,\hat v_0)_k}
\end{align}
where $x_0$ and $v_0$ are the atom coordinate and atom type of $M_0$, respectively\footnote{$\mathbf{x}_0$ and $x_0$ are different here. The former represents a general data sample in the classic diffusion process while the latter stands for the atom coordinate of the original molecule $M_0$}.
$\hat{x}_0$ and $\hat{v}_0$ are the predictions of the diffusion model $\phi$, given by
$\gamma_t=\frac{\bar\alpha_{t-1}\beta_t^2}{2\Sigma_\theta(x_t,t)(1-\bar\alpha_t)^2}$ and $C$ is a constant. Note that, we adopt $\phi_\theta(M^*_t,t,\mathcal{P})$ as the diffusion model and the denoising probability $p_\theta(M_{t-1}|M^*_t,\mathcal{P}))$ can be derived based on it.
\begin{equation}
\hat{M}_0 = [\hat{x}_0, \hat{v}_0] = \phi_\theta(M^*_t,t,\mathcal{P})
\end{equation}


\textbf{Pseudo molecule estimation.}\label{subsec:pme}
The main focus of this paper is to figure out an approximated sample representation $M^*_t$ that can bridge the gap between the training and inference of diffusion-based molecule generation. Therefore, $M^*_t$ should satisfy the following two properties. On the one hand, $M^*_t$ is expected to contain the key information of estimated pseudo molecule at the $t$-th step (i.e., $M_t$). On the other hand, $M^*_t$ must resemble the iterative output of the diffusion model. 

To this end, we propose pseudo molecule estimation, whose core idea is to regard the model's current predictions as the ground truth instead of using the standard noisy sample in training time. Using this estimate allows the model to make the data distribution during training a weighted average over the true distribution and the model's learned distribution, thus reducing the gap between training and inference.

Concretely, the pseudo molecule estimation leverages the diffusion model to re-predict the original molecule state based on the molecule state at step $t$. As a result, an estimated original molecule state is obtained.
\begin{align}
\widetilde M_0 = \phi_\theta(M_t,t,\mathcal{P})
\end{align}
Next, we apply Bayesian theory and add noises into the estimated original molecule $\widetilde M_0$, thus yielding a new variable $\widetilde M_t$ that describes the molecule state at step $t$.
\begin{align}
\widetilde M_t \sim q(\widetilde M_t|\widetilde M_0, \mathcal{P}) 
\end{align}

In this way, the estimated original molecule state $\widetilde M_0$ mimics the scenario in the inference stage where the true molecule state is not available. And $\widetilde M_t$ is a pseudo molecule state at step $t$, which can satisfy the two properties of $M^*_t$. However, we do not directly use the estimated molecule state $\widetilde M_0$ as  $M^*_t$, because purely adopting the model prediction as the condition would contaminate the training process. Instead, $M^*_t$ picks $M_t$ with probability $p_T$, otherwise chooses the estimated molecule $\widetilde M_{t}$. Hence, the training process of the diffusion model can be controlled by the pre-defined probability  $p_T$. 

Note that although the pseudo molecule estimation applies the diffusion model $\phi$ to produce the condition $M^*_t$, this process does not need to receive gradients and optimize. That is, the pseudo molecule estimation will dynamically use the diffusion model $\phi$ to add noises into the condition in the training stage and thus narrow the difference between the learning and inference.
\begin{equation}
\label{pme:eq:pick}
P(M^*_t) = 
\left\{\begin{matrix} 
  p_T,   & \;M^*_t = M_{t}  \\  
  1-p_T, & \;M^*_t = \widetilde M_{t}
\end{matrix}\right. 
\end{equation}


\textbf{Probability temperature annealing.}\label{subsec:pta}
Hoping the model can learn the data distribution more smoothly during training, we use monotonic decline functions to control the selection probability.
Intuitively, we need this probability curve to have a lower cooling rate at the beginning of training, that is, a smaller derivative value, and a higher cooling rate at the end of training, so that it can quickly adapt to changes from training to inference. We take the OR~\citep{zhang2019bridge} model's curve as the first one, which goes with a hyperparameter (equation~\ref{pta:eq:ori}) and is borrowed from Bengio~\citep{bengio2015scheduled}. However, under its default setting (shown in Setup), the probability $p_T$ cools too quickly, which is not conducive to learning a robust data distribution.

\begin{align}
p_T & = \frac{\mu}{\mu + {\rm exp}({e/\mu})} \label{pta:eq:ori}
\end{align}

To make the learning process more stable, we propose two other temperature anneals, one is linear annealing (equation~\ref{pta:eq:arc} left) and the other is arc annealing (equation~\ref{pta:eq:arc} right). For the sake of conciseness, this can also be regarded as $p = \sqrt{r^2 - e^2}/r$), we use a modified quarter circle curve as the cooling curve, make the result a real number and finally regularize it to ensure that the curve value range is in $[0, 1]$). In order to avoid $p_T$ being too low in the later stages of training, the model basically enters the self-verification learning stage, which has been reinforcing bias and making it difficult to learn the original distribution. We use ${\rm min}(p, {\rm lower\_bound})$ to obtain the probability of actual use.

\begin{align}
p_T = 1 + {\rm slope} * e
,\quad\quad\quad\quad\quad
p_T = \sqrt{{\rm max}(r^2 - (e/100)^2, 0)}/r \label{pta:eq:arc}
\end{align}

\section{Experiments}
\begin{table*}[ht]
\centering
\setlength{\tabcolsep}{0.5mm}
\caption{We compared our models with +Ours sequentially with all other models, indicating the best performing method in each case in \textbf{bold} and highlighting the metrics where our method achieved second place with \underline{underlining}. In the subsequent text, we will use \textsc{DiffGap} to represent BindDM+Ours.}
\footnotesize
\begin{tabular}{c|cccccccccccccc}
    \hline
    & \multicolumn{2}{c}{Vina Score($\downarrow$)} & \multicolumn{2}{c}{Vina Min($\downarrow$)}   & \multicolumn{2}{c}{Vina Dock($\downarrow$)}  & \multicolumn{2}{c}{High Affinity($\uparrow$)} & \multicolumn{2}{c}{QED($\uparrow$)} & \multicolumn{2}{c}{SA($\uparrow$)} & \multicolumn{2}{c}{Div($\uparrow$)} \\
    Models & Avg. & Med. & Avg. & Med. & Avg. & Med. & Avg. & Med. & Avg. & Med. & Avg. & Med. & Avg. & Med. \\
    \hline
    Reference  & -6.36 & -6.46 & 6.71 & -6.49 & -7.45 & -7.26 & - & - & 0.48 & 0.47 & 0.73 & 0.74 & - & - \\
    \hline
    liGAN        & - & - & -  & - & -6.33 & -6.20 & 21.1\% & 11.1\% & 0.39 & 0.39 & 0.59 & 0.57 & 0.66 & 0.67 \\
    GraphBP      & - & - & -  & - & -4.80 & -4.70 & 14.2\% & 6.7\%  & 0.43 & 0.45 & 0.49 & 0.48 & \textbf{0.79} & \textbf{0.78} \\
    AR           & -5.75 & -5.64 & -6.18 & -5.88 & -6.75 & -6.62 & 37.9\% & 31.0\% & 0.51 & 0.50 & 0.63 & 0.63 & 0.70 & 0.70\\
    Pocket2Mol   & -5.14 & -4.70 & -6.42 & -5.82 & -7.15 & -6.79 & 48.4\% & 51.1\% & \textbf{0.56} & \textbf{0.57} & \textbf{0.74} & \textbf{0.75} & 0.69 & 0.71 \\
    DecompDiff   & -5.67 & -6.04 & -7.04 & -7.09 & -8.39 & \underline{-8.43} & 64.4\% & 71.0\% & 0.45 & 0.43 & 0.61 & 0.60 & 0.68 & 0.68 \\
    MolCRAFT & \textbf{-6.59} & -7.04 & -7.27 & -7.26 & -7.92 & -8.01 & - & - & 0.50 & - & 0.69 & - & 0.72 & - \\
    \hline
    TargetDiff   & -5.47 & -6.30 & -6.64 & -6.83 & -7.80 & -7.91 & 58.1\% & 59.1\% & 0.48 & 0.48 & 0.58 & 0.58 & 0.72 & 0.71\\
    +Ours & \underline{-6.51} & \textbf{-7.18} & \textbf{-7.50} & \textbf{-7.38} & \textbf{-8.54} & \textbf{-8.38} & \underline{59.0\%} & \underline{62.8\%} & 0.46 & 0.46 & 0.56 & 0.56 & \textbf{0.79} & \underline{0.77} \\
    \hline
    BindDM       & -5.92 & -6.81 & -7.29 & -7.34 & -8.41 & -8.37 & 64.8\% & 71.6\% & 0.51 & 0.52 & 0.58 & 0.58 & 0.75 & 0.74\\
    +Ours (DiffGap)     & \underline{-6.28} & \underline{-6.90} & \textbf{-7.39} & \textbf{-7.45} & \textbf{-8.43} & \textbf{-8.47} & \textbf{68.9\%} & \textbf{72.2\%} & \underline{0.51} & \underline{0.52} & 0.59 & 0.58 & 0.75 & 0.75\\
    \hline
\end{tabular}
\label{exp:tab:vina}
\end{table*}

\subsection{Setup}


\textbf{Data.} We use CrossDocked2020~\citep{crossdock2020}, the commonly used protein-ligand pairs dataset, as a benchmark dataset for both training and evaluation. Similar to~\citep{luo20213d, targetdiff, guan2023decompdiff, huang2024binddm}, we filter the complexes with RMSD (Root Mean Square Deviation, the measure of the average distance between the atoms of superimposed molecules) higher than 1 $\mathring{\text{A}}$ and remains 100,000 pairs for training, 100 pairs for testing.

\textbf{Baseline.} We use our model to compare the affinity of the generated molecules with liGAN~\citep{liGAN}, AR~\citep{luo20213d}, Pocket2Mol~\citep{peng2022pocket2mol}, GraphBP~\citep{GraphBP}, TargetDiff~\citep{targetdiff}, DecompDiff~\citep{guan2023decompdiff} and BindDM~\citep{huang2024binddm}. In particular, 
TargetDiff, DecompDiff, and BindDM represent previous state-of-the-art performance in 3D molecule generation to a given protein structure with diffusion process and equivariant graph neural network, considering rotational and translational equivariance.

\textbf{DiffGap.} We employ the E(n)-Equivariant GNN~\citep{satorras2021En} as the backbone model, which contains 9 equivariant layers and makes diffusion steps $T=1000$ the same as DDPM. As liGAN and AR do, we choose OpenBabel~\citep{o2011open} to reconstruct the 3D molecule from atom coordinates.
We use the Adam optimizer, with $\beta=(0.95, 0.999)$ and a learning rate of 1e-4, without weight decay. We set the max training step to 200K and the batch size to 4 for all the models.
In the scenario, the probability of not computing the prediction $P(t=T-1)^{\rm batch size}$ is 0.996, which is acceptable and can be ignored.
More details can be found in the README file within the code repository.

\subsection{Results}


We conduct several experiments to compare our method against the aforementioned baseline models, primarily evaluating the performance of our approach in terms of the quality of generated molecules.
The main data are presented in four aspects: chemical bond distributions, binding affinity, and molecular properties of DiffGap. Additionally, we perform ablation experiments to verify the rationale behind the hyperparameter selection of the adaptive sampling strategy.

\begin{figure}[h]
    \centering
    \includegraphics[width=0.8\linewidth]{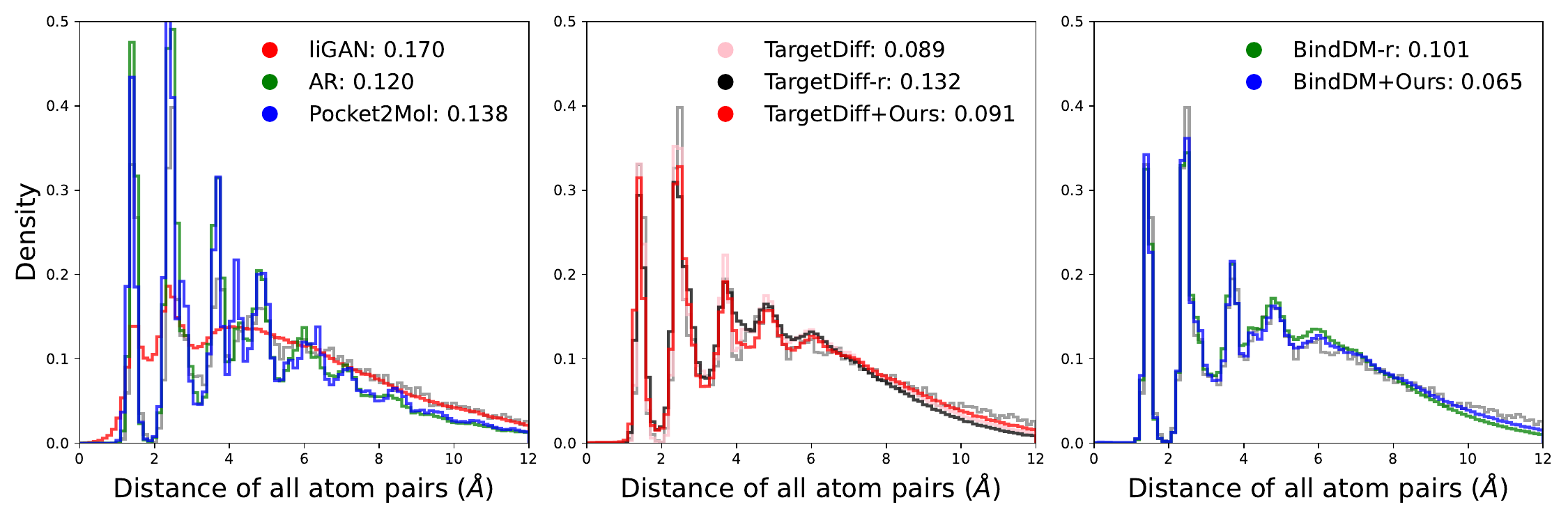}
    \caption{Comparing the distribution for distances of all-atom for reference molecules in the test set (gray) and generated molecules (color). Jensen-Shannon divergence (JSD$\downarrow$) between two distributions is reported.}
    \label{exp:fig:jsd}
\end{figure}

\textbf{Bond distribution.}
First, we consider the distribution of atomic coordinates and the most common types of bonds connected to carbon atoms after applying the adaptive sampling strategy.
In particular, the bond types include carbon-carbon bonds, carbon-nitrogen bonds, and carbon-oxygen bonds, covering single bonds(`-'), double bonds(`='), and aromatic bonds(`:'). We evaluate the difference between the generated bonds and the reference bonds using the Jensen-Shannon divergence~\citep{lin1991divergence}, where a lower value indicates better performance.

As shown in Table~\ref{exp:tab:jsd}, the molecules generated by our method are among the best three results, comparable to BindDM and surpassing other models. Meanwhile, our approach exhibits a significant advantage across all carbon bonds, particularly in double and aromatic bonds, where the latter achieves a twofold improvement over the previous best result.
We see in Figure~\ref{exp:fig:jsd} that \textsc{DiffGap} achieves the lowest JSD of 0.065 to reference in all-atom pairs distance distribution of the ligands in test sets, exceeding notably TargetDiff and others. Moreover, TargetDiff+Ours beats TargetDiff-r (the reproduced TargetDiff). That is to say, the molecules generated by \textsc{DiffGap} surpass those produced by prior methods overall.

\begin{table}[h]
\centering
\small
\caption{Jensen-Shannon divergence between the bond distance for reference and the generated. And we highlight the best 3 results with \textbf{bold text}, \underline{underlined text}, and \textit{italic text} respectively.}
\begin{tabular}{c|ccccccc}
  \hline
  Bond($\downarrow$) & liGAN & AR & Pocket2Mol & TargetDiff & DecompDiff & BindDM & \textsc{DiffGap}
  \\\hline
    C-C & 0.601 & 0.609 & 0.496 & \textit{0.369} & \underline{0.359} & 0.380 & \textbf{0.357}
  \\C-N & 0.634 & 0.474 & 0.416 & 0.363 & \textit{0.344} & \underline{0.265} & \textbf{0.253}
  \\C-O & 0.656 & 0.492 & 0.454 & 0.421 & \underline{0.376} & \textbf{0.329} & \textit{0.388}
  \\C=C & 0.665 & 0.620 & 0.561 & \textit{0.505} & 0.537 & \underline{0.229} & \textbf{0.189}
  \\C=N & 0.749 & 0.635 & 0.629 & \textit{0.550} & 0.584 & \textbf{0.245} & \underline{0.260}
  \\C=O & 0.661 & 0.558 & 0.516 & 0.461 & \underline{0.374} & \textbf{0.249} & \textit{0.376}
  \\C:C & 0.497 & 0.451 & 0.416 & \textit{0.263} & \underline{0.251} & 0.282 & \textbf{0.141}
  \\C:N & 0.638 & 0.552 & 0.487 & \underline{0.235} & 0.269 & \textbf{0.130} & \textit{0.240}
  \\\hline
\end{tabular}
\label{exp:tab:jsd}
\end{table}

\begin{figure*}[h]
  \centering
  \includegraphics[width=\linewidth]{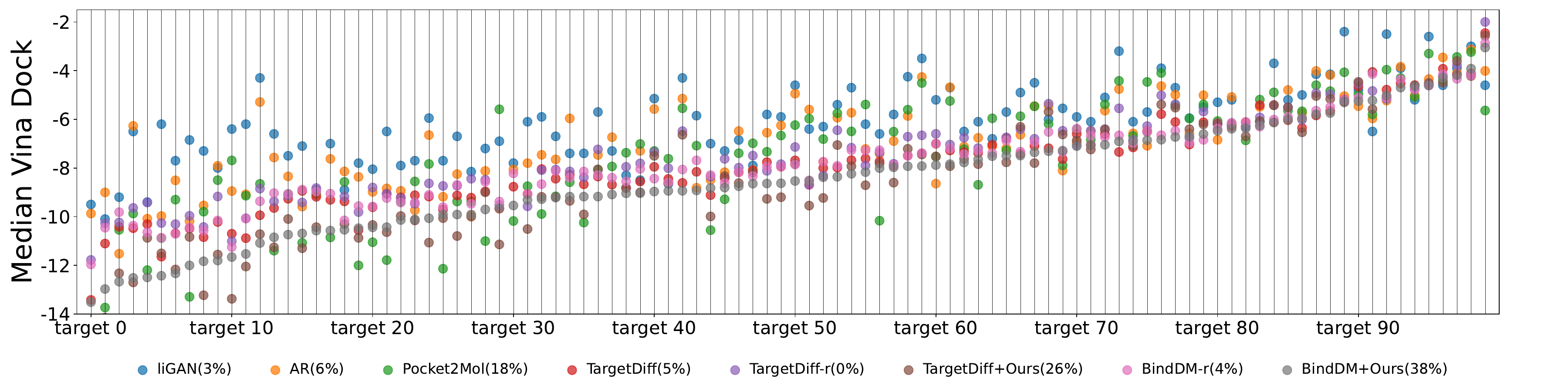}
  \caption{Median Vina dock energy for five models across 100 testing targets. The percentage represents the proportion of the model achieving the best binding affinity.}
  \label{exp:fig:vina}
\end{figure*}

\textbf{Binding affinity.}
The binding affinity between a molecule and a protein is measured by the energy released after binding.
AutoDock Vina~\citep{eberhardt2021autodock} is usually used to calculate the energy, which acts as a crucial evaluation metric.
Therefore, we use the docking scores and the results compared with the reference as metrics to compare and evaluate all the models,: Vina Score, Vina Min, Vina Dock, and High Affinity.
Vina Score is used to evaluate the stability of the small molecule-protein binding, Vina Min represents the minimum value during the docking process, Vina Dock energy attempts to find the lowest energy binding conformation, and High Affinity exhibits the proportion of superiority over the reference on Vina Dock.
As indicated in Table~\ref{exp:tab:vina}, \textsc{DiffGap} obtains higher mean and median in all affinity-related metrics compared to the baselines with the highest improvement reaching 4.1\% in the mean of High Affinity.
Otherwise, we reorder the median Vina Dock in ascending order according to our method for eight methods. 
As evidenced by Figure~\ref{exp:fig:vina}, DiffGap enables substantially superior performance compared to existing methods, yielding a minimum 20\% enhancement over TargetDiff and exceeding BindDM by more than 30\%.

\textbf{Molecular properties.} As for property-related metrics, drug-likeness QED~\citep{qed2012}, synthesizability SA~\citep{sa2009} and diversity are commonly used for evaluation.
Unlike that DecompDiff accepts a trade-off between property-related metrics and affinity-related metrics, we maintain proper properties and make somewhat progress in the mean of SA and the median of diversity in Table~\ref{exp:tab:vina}.
Nevertheless, we put less attention on QED and SA because many invalid molecules will be filtered out by virtual screening and it would be acceptable in a reasonable range.

\subsection{Ablation study}

These are two ablation studies to validate the rationality of our method.
The first study focuses on the selection of annealing strategies, while the second involves the specific parameter settings of the selected annealing method.

\textbf{Choice of annealing strategy.}
We employ ablation experiments for comparison to filter out the best method, which considers the impact of both the lower bound and the annealing method. Due to the time-consuming nature of sampling and evaluating the full test dataset of 100 targets, we randomly selected 10 targets for the ablation experiments. Then we used the docking scores as the evaluation metrics and calculated the time cost multiplier of DiffGap compared to the classic way (marked as Cost in Table~\ref{exp:tab:anneal}).





Overall, arc annealing outperformed the others at both lower bounds of 0.5 and 0.8, with it performing best when the lower bound was set to 0.5 in Table~\ref{exp:tab:anneal}. Consequently, we ultimately choose arc annealing and set the lower bound to 0.5, and we think the extra time cost is acceptable.


\begin{table}[ht]
\centering
\setlength{\tabcolsep}{1mm}
\small
\caption{Annealing methods comparison across 10 testing targets.}
\begin{tabular}{c|ccccccccc}
    \hline
    Metrics & Lower & \multicolumn{2}{c}{Vina Score($\downarrow$)} & \multicolumn{2}{c}{Vina Min($\downarrow$)}   & \multicolumn{2}{c}{Vina Dock($\downarrow$)} &  & Cost($\downarrow$)\\
    \diagbox{Annealing}{} & Bound & Avg. & Med. & Avg. & Med. & Avg. & Med. & Avg. & Avg. \\ \hline
    Original (equation~\ref{pta:eq:ori}) & 0.5 & -4.588 & -4.705 & -4.898 & -4.893 & -5.661 & -5.540 & -5.047 & 1.38 \\
     & 0.8 & -4.209 & -4.343 & -4.544 & -4.580 & -5.471 & -5.392 & -4.757 & 1.18 \\ \hline
    Linear (equation~\ref{pta:eq:arc} left) & 0.5 & -6.564 & \textbf{-6.499} & -6.617 & -6.241 & -7.421 & -7.249 & -6.765 & 1.37 \\
    ~ & 0.8 & \textbf{-6.662} & -6.418 & -6.593 & -6.223 & -7.317 & -7.146 & -6.727 & 1.18 \\ \hline
    Arc (equation~\ref{pta:eq:arc} right) & 0.5 & -6.496 & -6.384 & \textbf{-6.780} & \textbf{-6.525} & \textbf{-7.470} & \textbf{-7.369} & \textbf{-6.837} & 1.19 \\
    ~ & 0.8 & -6.413 & -6.189 & -6.719 & -6.445 & -7.463 & -7.398 & -6.771 & \textbf{1.12} \\ \hline

\end{tabular}
\label{exp:tab:anneal}
\end{table}

\textbf{Arc annealing comparison.}
We initially hypothesized that the arc annealing curve would yield the best results when $r=2$, although it remains to be validated. Therefore, we design a series of ablation experiments, selecting the values 1.5, 2, 3, 4, 8, and infinity to empirically determine the optimal value for the hyperparameter $r$. The experimental outcomes confirm our initial hypothesis. 

When $r$ equals infinity, $p_T$ is 1, which corresponds to not using our method at all, resulting in the lowest Vina scores, thereby proving the effectiveness of our method. When $r$ is set to 1.5, 3, 4, or 8, the results are comparable, indicating that even with the early introduction of estimation or fewer opportunities for estimation, there is still a certain degree of improvement. When $r$ equals 2, the best results are achieved, surpassing TargetDiff.

\begin{table}[ht]
\caption{Annealing comparison for better hyper-parameter.}
\centering
\begin{tabular}{c|ccccccc}
    \hline
    Metrics & \multicolumn{2}{c}{Vina Score($\downarrow$)} & \multicolumn{2}{c}{Vina Min($\downarrow$)}   & \multicolumn{2}{c}{Vina Dock($\downarrow$)} & Cost($\downarrow$)\\
    \diagbox{$r$}{} & Avg.           & Med.           & Avg.           & Med.           & Avg.           & Med. & Avg. \\ \hline
  
    $\infty$ & -5.41          & -6.01          & -6.32          & -6.28          & -7.28          & -7.39 & 1 \\ 
    8  & -5.55          & -6.27          & -6.49          & -6.54          & -7.35          & -7.51 & 1.01\\
    4       & -5.58 & -6.10 & -6.49 & -6.54 & -7.35 & -7.51 & 1.04\\
    3 & -5.61 & -6.27 & -6.49 & -6.48 & -7.53 & -7.53 & 1.08\\
    2       & \textbf{-6.51} & \textbf{-7.18} & \textbf{-7.50} & \textbf{-7.38} & \textbf{-8.54} & \textbf{-8.38} & 1.12\\
    1.5  & -5.63          & -6.12          & -6.41          & -6.37          & -7.32          & -7.42 & 1.14\\
       \hline
\end{tabular}
\label{exp:tab:rng}
\end{table}

\section{Conclusion}
DiffGap addresses critical limitations in diffusion-based molecular generation through adaptive sampling and pseudo-molecule estimation. By dynamically aligning training trajectories with inference dynamics to mitigate exposure bias and error accumulation, the framework ensures stable learning of 3D molecular distributions while preemptively adapting to input biases.Evaluations on CrossDocked2020 confirm DiffGap’s superiority in generating molecules with enhanced docking scores and binding affinities, outperforming existing methods.
The primary impact of this work is its versatility as a general-purpose framework that can be integrated as a plug-in to enhance various diffusion-based architectures. By generating molecules with superior docking scores and binding affinities, DiffGap offers a robust computational toolkit to accelerate structure-driven drug discovery.
However, we acknowledge that this mechanism is specific to the diffusion model and that the underlying principles are not yet understood, making it difficult to continue optimizing.

\acks{This work was supported by the National Natural Science Foundation of China (No. 62206192), the Natural Science Foundation of Sichuan Province (No. 2023NSFSC1408, 2024NSFTD0048), and the Science and Technology Major Project of Sichuan Province (No. 2024ZDZX0003). We also acknowledge the support of Sichuan Province Engineering Technology Research Center of Broadband Electronics Intelligent Manufacturing.}

\bibliography{acml25}

@article{ddpm,
  title={Denoising diffusion probabilistic models},
  author={Ho, Jonathan and Jain, Ajay and Abbeel, Pieter},
  journal={Advances in neural information processing systems},
  volume={33},
  pages={6840--6851},
  year={2020}
}

@inproceedings{yang2018unbiased,
  title={Unbiased offline recommender evaluation for missing-not-at-random implicit feedback},
  author={Yang, Longqi and Cui, Yin and Xuan, Yuan and Wang, Chenyang and Belongie, Serge and Estrin, Deborah},
  booktitle={Proceedings of the 12th ACM conference on recommender systems},
  pages={279--287},
  year={2018}
}

@inproceedings{ning2023input,
  title={Input Perturbation Reduces Exposure Bias in Diffusion Models},
  author={Ning, Mang and Sangineto, Enver and Porrello, Angelo and Calderara, Simone and Cucchiara, Rita},
  booktitle={International Conference on Machine Learning},
  pages={26245--26265},
  year={2023},
  organization={PMLR}
}

@inproceedings{targetdiff,
  title={3D Equivariant Diffusion for Target-Aware Molecule Generation and Affinity Prediction},
  author={Guan, Jiaqi and Qian, Wesley Wei and Peng, Xingang and Su, Yufeng and Peng, Jian and Ma, Jianzhu},
  booktitle={The Eleventh International Conference on Learning Representations},
  year={2022}
}

@article{luo20213d,
  title={A 3D generative model for structure-based drug design},
  author={Luo, Shitong and Guan, Jiaqi and Ma, Jianzhu and Peng, Jian},
  journal={Advances in Neural Information Processing Systems},
  volume={34},
  pages={6229--6239},
  year={2021}
}

@inproceedings{peng2022pocket2mol,
  title={Pocket2mol: Efficient molecular sampling based on 3d protein pockets},
  author={Peng, Xingang and Luo, Shitong and Guan, Jiaqi and Xie, Qi and Peng, Jian and Ma, Jianzhu},
  booktitle={International Conference on Machine Learning},
  pages={17644--17655},
  year={2022},
  organization={PMLR}
}

@inproceedings{zhang2019bridge,
    title = "Bridging the Gap between Training and Inference for Neural Machine Translation",
    author = "Zhang, Wen  and
      Feng, Yang  and
      Meng, Fandong  and
      You, Di  and
      Liu, Qun",
    booktitle = "Proceedings of the 57th Annual Meeting of the Association for Computational Linguistics",
    month = jul,
    year = "2019",
    pages = "4334--4343",
}

@article{bengio2015scheduled,
  title={Scheduled sampling for sequence prediction with recurrent neural networks},
  author={Bengio, Samy and Vinyals, Oriol and Jaitly, Navdeep and Shazeer, Noam},
  journal={Advances in neural information processing systems},
  volume={28},
  year={2015}
}

@article{lin1991divergence,
  title={Divergence measures based on the Shannon entropy},
  author={Lin, Jianhua},
  journal={IEEE Transactions on Information theory},
  volume={37},
  number={1},
  pages={145--151},
  year={1991},
  publisher={IEEE}
}

@inproceedings{sohl2015dpm,
  title={Deep unsupervised learning using nonequilibrium thermodynamics},
  author={Sohl-Dickstein, Jascha and Weiss, Eric and Maheswaranathan, Niru and Ganguli, Surya},
  booktitle={International conference on machine learning},
  pages={2256--2265},
  year={2015},
  organization={PMLR}
}

@article{crossdock2020,
  title={Three-dimensional convolutional neural networks and a cross-docked data set for structure-based drug design},
  author={Francoeur, Paul G and Masuda, Tomohide and Sunseri, Jocelyn and Jia, Andrew and Iovanisci, Richard B and Snyder, Ian and Koes, David R},
  journal={Journal of chemical information and modeling},
  volume={60},
  number={9},
  pages={4200--4215},
  year={2020},
  publisher={ACS Publications}
}

@article{liGAN,
  title={Generating 3D molecules conditional on receptor binding sites with deep generative models},
  author={Ragoza, Matthew and Masuda, Tomohide and Koes, David Ryan},
  journal={Chemical science},
  volume={13},
  number={9},
  pages={2701--2713},
  year={2022},
  publisher={Royal Society of Chemistry}
}

@inproceedings{GraphBP,
  title={Generating 3D Molecules for Target Protein Binding},
  author={Liu, Meng and Luo, Youzhi and Uchino, Kanji and Maruhashi, Koji and Ji, Shuiwang},
  booktitle={International Conference on Machine Learning},
  pages={13912--13924},
  year={2022},
  organization={PMLR}
}

@article{fuchs2020se,
  title={Se (3)-transformers: 3d roto-translation equivariant attention networks},
  author={Fuchs, Fabian and Worrall, Daniel and Fischer, Volker and Welling, Max},
  journal={Advances in neural information processing systems},
  volume={33},
  pages={1970--1981},
  year={2020}
}

@inproceedings{satorras2021En,
  title={E (n) equivariant graph neural networks},
  author={Satorras, V{\i}ctor Garcia and Hoogeboom, Emiel and Welling, Max},
  booktitle={International conference on machine learning},
  pages={9323--9332},
  year={2021},
  organization={PMLR}
}

@article{o2011open,
  title={Open Babel: An open chemical toolbox},
  author={O'Boyle, Noel M and Banck, Michael and James, Craig A and Morley, Chris and Vandermeersch, Tim and Hutchison, Geoffrey R},
  journal={Journal of cheminformatics},
  volume={3},
  pages={1--14},
  year={2011},
  publisher={Springer}
}

@article{eberhardt2021autodock,
  title={AutoDock Vina 1.2. 0: New docking methods, expanded force field, and python bindings},
  author={Eberhardt, Jerome and Santos-Martins, Diogo and Tillack, Andreas F and Forli, Stefano},
  journal={Journal of chemical information and modeling},
  volume={61},
  number={8},
  pages={3891--3898},
  year={2021},
  publisher={ACS Publications}
}

@article{qed2012,
  title={Quantifying the chemical beauty of drugs},
  author={Bickerton, G Richard and Paolini, Gaia V and Besnard, J{\'e}r{\'e}my and Muresan, Sorel and Hopkins, Andrew L},
  journal={Nature chemistry},
  volume={4},
  number={2},
  pages={90--98},
  year={2012},
  publisher={Nature Publishing Group}
}

@article{sa2009,
  title={Estimation of synthetic accessibility score of drug-like molecules based on molecular complexity and fragment contributions},
  author={Ertl, Peter and Schuffenhauer, Ansgar},
  journal={Journal of cheminformatics},
  volume={1},
  pages={1--11},
  year={2009},
  publisher={Springer}
}

@book{sneader2005drug,
  title={Drug discovery: a history},
  author={Sneader, Walter},
  year={2005},
  publisher={John Wiley \& Sons}
}

@article{mandal2009rational,
  title={Rational drug design},
  author={Mandal, Soma and Mandal, Sanat K and others},
  journal={European journal of pharmacology},
  volume={625},
  number={1-3},
  pages={90--100},
  year={2009},
  publisher={Elsevier}
}

@article{batool2019structure,
  title={A structure-based drug discovery paradigm},
  author={Batool, Maria and Ahmad, Bilal and Choi, Sangdun},
  journal={International journal of molecular sciences},
  volume={20},
  number={11},
  pages={2783},
  year={2019},
  publisher={MDPI}
}

@article{mayr2009novel,
  title={Novel trends in high-throughput screening},
  author={Mayr, Lorenz M and Bojanic, Dejan},
  journal={Current opinion in pharmacology},
  volume={9},
  number={5},
  pages={580--588},
  year={2009},
  publisher={Elsevier}
}

@article{zhang2022molecular,
  title={Molecular docking-based computational platform for high-throughput virtual screening},
  author={Zhang, Baohua and Li, Hui and Yu, Kunqian and Jin, Zhong},
  journal={CCF Transactions on High Performance Computing},
  pages={1--12},
  year={2022},
  publisher={Springer}
}

@article{alphafold,
  title={Highly accurate protein structure prediction with AlphaFold},
  author={Jumper, John and Evans, Richard and Pritzel, Alexander and Green, Tim and Figurnov, Michael and Ronneberger, Olaf and Tunyasuvunakool, Kathryn and Bates, Russ and {\v{Z}}{\'\i}dek, Augustin and Potapenko, Anna and others},
  journal={nature},
  volume={596},
  number={7873},
  pages={583--589},
  year={2021},
  publisher={Nature Publishing Group}
}

@article{alphafold3,
  title={Accurate structure prediction of biomolecular interactions with AlphaFold 3},
  author={Abramson, Josh and Adler, Jonas and Dunger, Jack and Evans, Richard and Green, Tim and Pritzel, Alexander and Ronneberger, Olaf and Willmore, Lindsay and Ballard, Andrew J and Bambrick, Joshua and others},
  journal={Nature},
  pages={1--3},
  year={2024},
  publisher={Nature Publishing Group UK London}
}

@inproceedings{pemasinghe2021simulated,
  title={Simulated Annealing and It’s Application in Molecular Structure Optimizations},
  author={Pemasinghe, Sajeewa and Abeygunawardhana, Pradeep KW},
  booktitle={2021 10th International Conference on Information and Automation for Sustainability (ICIAfS)},
  pages={258--262},
  year={2021},
  organization={IEEE}
}

@article{you2018graph,
  title={Graph convolutional policy network for goal-directed molecular graph generation},
  author={You, Jiaxuan and Liu, Bowen and Ying, Zhitao and Pande, Vijay and Leskovec, Jure},
  journal={Advances in neural information processing systems},
  volume={31},
  year={2018}
}

@article{jensen2019graph,
  title={A graph-based genetic algorithm and generative model/Monte Carlo tree search for the exploration of chemical space},
  author={Jensen, Jan H},
  journal={Chemical science},
  volume={10},
  number={12},
  pages={3567--3572},
  year={2019},
  publisher={Royal Society of Chemistry}
}

@article{paassen2022recursive,
  title={Recursive tree grammar autoencoders},
  author={Paa{\ss}en, Benjamin and Koprinska, Irena and Yacef, Kalina},
  journal={Machine Learning},
  volume={111},
  number={9},
  pages={3393--3423},
  year={2022},
  publisher={Springer}
}

@article{grisoni2020bidirectional,
  title={Bidirectional molecule generation with recurrent neural networks},
  author={Grisoni, Francesca and Moret, Michael and Lingwood, Robin and Schneider, Gisbert},
  journal={Journal of chemical information and modeling},
  volume={60},
  number={3},
  pages={1175--1183},
  year={2020},
  publisher={ACS Publications}
}

@article{arus2019randomized,
  title={Randomized SMILES strings improve the quality of molecular generative models},
  author={Ar{\'u}s-Pous, Josep and Johansson, Simon Viet and Prykhodko, Oleksii and Bjerrum, Esben Jannik and Tyrchan, Christian and Reymond, Jean-Louis and Chen, Hongming and Engkvist, Ola},
  journal={Journal of cheminformatics},
  volume={11},
  pages={1--13},
  year={2019},
  publisher={Springer}
}

@article{schoenmaker2023uncorrupt,
  title={UnCorrupt SMILES: a novel approach to de novo design},
  author={Schoenmaker, Linde and B{\'e}quignon, Olivier JM and Jespers, Willem and van Westen, Gerard JP},
  journal={Journal of Cheminformatics},
  volume={15},
  number={1},
  pages={22},
  year={2023},
  publisher={Springer}
}

@article{zeng2024chatmol,
  title={ChatMol: interactive molecular discovery with natural language},
  author={Zeng, Zheni and Yin, Bangchen and Wang, Shipeng and Liu, Jiarui and Yang, Cheng and Yao, Haishen and Sun, Xingzhi and Sun, Maosong and Xie, Guotong and Liu, Zhiyuan},
  journal={Bioinformatics},
  volume={40},
  number={9},
  pages={btae534},
  year={2024},
  publisher={Oxford University Press}
}

@inproceedings{brahmavar2024generating,
  title={Generating Novel Leads for Drug Discovery using LLMs with Logical Feedback},
  author={Brahmavar, Shreyas Bhat and Srinivasan, Ashwin and Dash, Tirtharaj and Krishnan, Sowmya Ramaswamy and Vig, Lovekesh and Roy, Arijit and Aduri, Raviprasad},
  booktitle={Proceedings of the AAAI Conference on Artificial Intelligence},
  volume={38},
  pages={21--29},
  year={2024}
}

@article{walters2020applications,
  title={Applications of deep learning in molecule generation and molecular property prediction},
  author={Walters, W Patrick and Barzilay, Regina},
  journal={Accounts of chemical research},
  volume={54},
  number={2},
  pages={263--270},
  year={2020},
  publisher={ACS Publications}
}

@article{lim2020scaffold,
  title={Scaffold-based molecular design with a graph generative model},
  author={Lim, Jaechang and Hwang, Sang-Yeon and Moon, Seokhyun and Kim, Seungsu and Kim, Woo Youn},
  journal={Chemical science},
  volume={11},
  number={4},
  pages={1153--1164},
  year={2020},
  publisher={Royal Society of Chemistry}
}

@inproceedings{jin2018junction,
  title={Junction tree variational autoencoder for molecular graph generation},
  author={Jin, Wengong and Barzilay, Regina and Jaakkola, Tommi},
  booktitle={International conference on machine learning},
  pages={2323--2332},
  year={2018},
  organization={PMLR}
}

@inproceedings{
    song2020score,
    title={Score-Based Generative Modeling through Stochastic Differential Equations},
    author={Yang Song and Jascha Sohl-Dickstein and Diederik P Kingma and Abhishek Kumar and Stefano Ermon and Ben Poole},
    booktitle={International Conference on Learning Representations},
    year={2021},
    url={https://openreview.net/forum?id=PxTIG12RRHS}
}

@inproceedings{guan2023decompdiff,
  title={DecompDiff: Diffusion Models with Decomposed Priors for Structure-Based Drug Design},
  author={Guan, Jiaqi and Zhou, Xiangxin and Yang, Yuwei and Bao, Yu and Peng, Jian and Ma, Jianzhu and Liu, Qiang and Wang, Liang and Gu, Quanquan},
  booktitle={International Conference on Machine Learning},
  pages={11827--11846},
  year={2023},
  organization={PMLR}
}

@inproceedings{huang2024binddm,
  title={Binding-Adaptive Diffusion Models for Structure-Based Drug Design},
  author={Huang, Zhilin and Yang, Ling and Zhang, Zaixi and Zhou, Xiangxin and Bao, Yu and Zheng, Xiawu and Yang, Yuwei and Wang, Yu and Yang, Wenming},
  booktitle={Proceedings of the AAAI Conference on Artificial Intelligence},
  volume={38},
  pages={12671--12679},
  year={2024}
}

@inproceedings{nichol2021glide,
  title={GLIDE: Towards Photorealistic Image Generation and Editing with Text-Guided Diffusion Models},
  author={Nichol, Alexander Quinn and Dhariwal, Prafulla and Ramesh, Aditya and Shyam, Pranav and Mishkin, Pamela and Mcgrew, Bob and Sutskever, Ilya and Chen, Mark},
  booktitle={International Conference on Machine Learning},
  pages={16784--16804},
  year={2022},
  organization={PMLR}
}

@article{lamb2016professor,
  title={Professor forcing: A new algorithm for training recurrent networks},
  author={Lamb, Alex M and ALIAS PARTH GOYAL, Anirudh Goyal and Zhang, Ying and Zhang, Saizheng and Courville, Aaron C and Bengio, Yoshua},
  journal={Advances in neural information processing systems},
  volume={29},
  year={2016}
}

@inproceedings{
ning2023elucidating,
title={Elucidating the Exposure Bias in Diffusion Models},
author={Mang Ning and Mingxiao Li and Jianlin Su and Albert Ali Salah and Itir Onal Ertugrul},
booktitle={The Twelfth International Conference on Learning Representations},
year={2024},
url={https://openreview.net/forum?id=xEJMoj1SpX}
}

@article{weininger1988smiles,
  title={SMILES, a chemical language and information system. 1. Introduction to methodology and encoding rules},
  author={Weininger, David},
  journal={Journal of chemical information and computer sciences},
  volume={28},
  number={1},
  pages={31--36},
  year={1988},
  publisher={ACS Publications}
}

@article{bohacek1996art,
  title={The art and practice of structure-based drug design: a molecular modeling perspective},
  author={Bohacek, Regine S and McMartin, Colin and Guida, Wayne C},
  journal={Medicinal research reviews},
  volume={16},
  number={1},
  pages={3--50},
  year={1996},
  publisher={Wiley Subscription Services, Inc., A Wiley Company New York}
}

@article{xue2025targetsa,
  title={Target{SA}: adaptive simulated annealing for target-specific drug design},
  author={Xue, Zhe and Sun, Chenwei and Zheng, Wenhao and Lv, Jiancheng and Liu, Xianggen},
  journal={Bioinformatics},
  volume={41},
  number={1},
  pages={btae730},
  year={2025},
  publisher={Oxford University Press}
}

@article{liu2021deep,
  title={Deep geometric representations for modeling effects of mutations on protein-protein binding affinity},
  author={Liu, Xianggen and Luo, Yunan and Li, Pengyong and Song, Sen and Peng, Jian},
  journal={PLoS computational biology},
  volume={17},
  number={8},
  pages={e1009284},
  year={2021},
  publisher={Public Library of Science San Francisco, CA USA}
}

\end{document}